%% file: main.tex
\documentclass[10pt,journal,compsoc]{IEEEtran}
\newif\ifpeerreview

\peerreviewfalse
\usepackage[nocompress]{cite}
\usepackage{url}
\usepackage{amsmath,amssymb,graphicx}
\usepackage{lipsum} 
\usepackage[switch]{lineno}

\newcommand{\paperID}{67}

\input{preambles}

\newcommand{\papertitle}{Shape from Polarization of\\Thermal Emission and Reflection}
\title{\papertitle}

\author{Kazuma~Kitazawa,
        and~Tsuyoshi~Takatani,~\IEEEmembership{Member,~IEEE}
        
\IEEEcompsocitemizethanks{
\IEEEcompsocthanksitem 
K. Kitazawa (kitazawa.kazuma.qy@alumni.tsukuba.ac.jp) and \\
T. Takatani (takatani@iit.tsukuba.ac.jp) are with \\University of Tsukuba, Japan.
}
}

\newcommand{\insertfig}{
    \centering
    \includegraphics[width=1.0\linewidth]{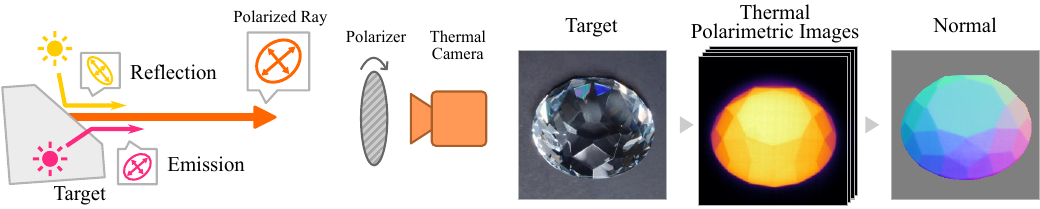}
    \captionof{figure}{
        Concept of the proposed method for estimating normals from thermal polarimetric images. Left: Our novel observation model showing how the polarization of radiation from a thermal object consists of two components: emission from the object and reflection from the environment. Right: An example of a transparent object with its visible light photograph, thermal polarimetric images captured by our prototype system, and the resulting estimated normal map.
    }
    \label{fig: concept}
}
\makeatletter
\apptocmd{\@maketitle}{%
 \setcounter{figure}{0}%
 \centering\insertfig%
}{}{}
\makeatother

\begin{document}

\IEEEtitleabstractindextext{%
\begin{abstract}
Shape estimation for transparent objects is challenging due to their complex light transport. To circumvent these difficulties, we leverage the Shape from Polarization (SfP) technique in the Long-Wave Infrared (LWIR) spectrum, where most materials are opaque and emissive. While a few prior studies have explored LWIR SfP, these attempts suffered from significant errors due to inadequate polarimetric modeling, particularly the neglect of reflection. To address this gap, we formulated a polarization model that explicitly accounts for the combined effects of emission and reflection. Based on this model, we estimated surface normals using not only a direct model-based method but also a learning-based approach employing a neural network trained on a physically-grounded synthetic dataset. Furthermore, we modeled the LWIR polarimetric imaging process, accounting for inherent systematic errors to ensure accurate polarimetry. We implemented a prototype system and created \thermalpol, the first real-world benchmark dataset for LWIR SfP. Through comprehensive experiments, we demonstrated the high accuracy and broad applicability of our method across various materials, including those transparent in the visible spectrum.

\end{abstract}

}

\maketitle


\input{01_intro}

\input{02_related_work}

\input{03_principle}
\input{04_method}

\input{05_implementation}
\input{06_experiment}

\input{07_conclusion}

\bibliographystyle{IEEEtran}
\bibliography{references}

\ifpeerreview \else


\section*{Acknowledgments}
This work was supported by JSPS KAKENHI Grant Numbers 22K18420.


\begin{IEEEbiographynophoto}{Kazuma~Kitazawa}
received his B.E. degree from University of Tsukuba in 2024.
He is currently pursuing an M.E. degree at University of Tsukuba,
focusing on computational imaging.
\end{IEEEbiographynophoto}

\begin{IEEEbiographynophoto}{Tsuyoshi~Takatani}
received the doctoral degree from the Nara Institute of Science and Technology (NAIST), in 2019. He is currently an assistant professor at the Institute of Systems and Information Engineering, University of Tsukuba. He leads the Computational Imaging and Graphics Laboratory as the founding director. His research interests include computational imaging and fabrication, and inverse rendering. He is a member of IEEE and OPTICA.
\end{IEEEbiographynophoto}

\vfill

\fi

\end{document}

%% file: preambles.tex
\usepackage[utf8]{inputenc}
\usepackage{amsmath}
\usepackage{amsfonts}
\usepackage{graphicx}
\usepackage{svg}
\usepackage{subfigure}
\usepackage{float}
\usepackage{xspace}
\usepackage{geometry}
\usepackage{bm}
\usepackage{siunitx}
\usepackage{mathtools}
\usepackage{caption}
\usepackage{subcaption}
\usepackage{physics}
\usepackage{tabularx}
\usepackage{booktabs}
\usepackage{xparse}

\usepackage[pdftex]{hyperref}
\hypersetup{
    colorlinks=true,
    urlcolor=blue,
    citecolor=blue,
    linkcolor=blue
}
\usepackage[nameinlink]{cleveref}
\crefname{figure}{Fig.}{Figs.}
\crefname{equation}{Eq.}{Eqs.}
\crefname{section}{Sec.}{Secs.}
\crefname{table}{TABLE}{TABLE}

\makeatletter
\DeclareRobustCommand\onedot{\futurelet\@let@token\@onedot}
\def\@onedot{\ifx\@let@token.\else.\null\fi\xspace}

\def\eg{\emph{e.g}\onedot}

\def\etal{\emph{et al}\onedot}
\makeatother

\DeclareSIUnit{\rad}{rad}

\newcommand{\jp}[1]{%
}

\ExplSyntaxOn
\NewDocumentCommand{\tablerow}{m}
 {
  \clist_set:Nn \l_tmpa_clist { #1 }
  \clist_use:Nn \l_tmpa_clist { & }
  \\
 }
\ExplSyntaxOff

\newcommand{\tbf}[1]{\textbf{#1}}

\renewcommand{\paragraph}[1]{\vspace{1ex}\noindent\textbf{#1}\quad}

\newcommand{\thermalpol}{\textit{ThermoPol}\xspace}
\newcommand{\thermalpolsixteen}{\textit{ThermoPol16}\xspace}

\newcommand{\boar}{\textit{\#3\,boar}\xspace}

\newcommand{\blackdog}{\textit{\#8\,blackdog}\xspace}

\newcommand{\garlic}{\textit{\#10\,garlic}\xspace}
\newcommand{\fish}{\textit{\#11\,fish}\xspace}

\newcommand{\monk}{\textit{\#16\,monk}\xspace}

\newcommand{\polangle}{\psi}
\newcommand{\iraw}{\mathit{I}\xspace}
\newcommand{\gainvec}{\mathbf{c}\xspace}
\newcommand{\gain}{\mathit{c}\xspace}
\newcommand{\mpol}{\mathbf{M}_\mathrm{pol}\xspace}
\newcommand{\mcam}{\mathbf{M}_\mathrm{cam}\xspace}
\newcommand{\mcampsi}{m\xspace}
\newcommand{\ext}{\mathbf{s}\xspace}

\newcommand{\offset}{\mathit{I}_\mathrm{off} (\polangle)\xspace}
\newcommand{\dolp}{\rho}
\newcommand{\aolp}{\varphi}
\newcommand{\zenith}{\theta}
\newcommand{\azimuth}{\phi}

\newcommand{\stefan}{\sigma}

\newcommand{\thetair}{{\theta}_R}
\newcommand{\thetait}{{\theta}_T}
\newcommand{\refp}{R_p}
\newcommand{\refs}{R_s}
\newcommand{\transp}{T_p}
\newcommand{\transs}{T_s}
\newcommand{\iref}{L_{\mathrm{R}}}
\newcommand{\itrans}{L_{\mathrm{E}}}
\newcommand{\out}{L}
\newcommand{\temp}{\tau}

%% file: 01_intro.tex
\IEEEraisesectionheading{
  \section{Introduction}
  \label{sec:intro}
}


\IEEEPARstart{S}{hape} estimation of transparent objects remains one of the most challenging problems in computer vision, despite being essential for applications ranging from industrial inspection to augmented reality. While surface normal estimation techniques have advanced significantly for opaque objects, they prove inadequate when faced with complex light transport in transparent materials like glass and plastic.

To overcome this fundamental challenge, leveraging Long-Wave Infrared (LWIR) imaging, also known as thermal imaging, has emerged as a promising direction~\cite{eren2009scanning, tanaka2019time, narayanan2024shape}. Many materials that are transparent in the visible spectrum (\eg, glass, acrylic) are opaque to the LWIR spectrum, making LWIR imaging particularly suitable for measuring their surfaces. Furthermore, since every surface acts as an emitter through thermal radiation, with all objects intrinsically radiating energy according to their temperature, LWIR imaging enables passive shape acquisition~\cite{nagase2022shape}.

Shape from Polarization (SfP) also offers a promising foundation for this work, as it infers surface normals from the polarization of light, enabling passive and detailed shape acquisition. While various SfP methods in the visible spectrum exist, including model-based~\cite{miyazaki2003polarization, mahmoud2012direct, zhao2016multi} and learning-based approaches~\cite{kondo2020accurate, ba2020deep, lei2022shape, shao2023transparent}, they generally struggle with transparent objects. In contrast, this work achieves SfP for transparent objects by utilizing LWIR imaging.
In fact, a few prior studies on LWIR SfP have been explored~\cite{partridge1995three, miyazaki2002determining, kechiche2017use}. However, their results suffered from significant errors due to inadequate modeling of the polarization of thermal radiation, particularly by neglecting reflection. Crucially, in the LWIR spectrum, the surrounding environment is also emissive, making radiation reflected from an object a non-negligible component.

In this paper, we propose a novel method to LWIR SfP based on the polarization state of thermal emission and reflection, as illustrated in \cref{fig: concept}. 
First, we formulate a polarization model that explicitly accounts for the combined effect of two components: emission from the surface (derived from the object) and reflection on the surface (derived from the environment). Second, we construct a model of the LWIR polarimetric imaging process that addresses realistic effects from stray LWIR light, the camera's polarimetric characteristics, and thermal noise. Third, we propose a physics-based learning method to estimate surface normals. The polarization model allows us to synthesize high-quality observations with a variety of shapes and materials in different environments. Training with this synthetic dataset enables accurate normal estimation of objects with complicated shapes and transparent materials.

\begin{figure}[tb]
    \centerline{\includegraphics[width=1.0\columnwidth]{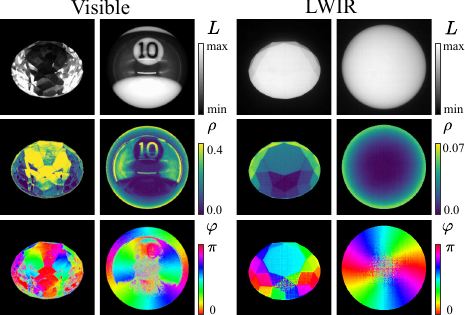}}
    \caption{
        Comparison of the polarization states in the visible and LWIR images. From top to bottom: radiance $L$, DoLP $\dolp$, and AoLP $\aolp$. The polarization state in the visible spectrum is significantly affected by transparency and surface albedos, compared to the LWIR spectrum.
    }
    \label{fig: lwir_vs_visible}
\end{figure}

A key feature of our method is the availability to not only transparent materials but also translucent or opaque materials with spatially varying albedos, due to spatially uniform emissivity and reflectance of an object. As shown in \cref{fig: lwir_vs_visible}, the radiance and the polarization state are less affected by the albedos.

Our main contributions are summarized as follows:
(1)~We propose a novel LWIR polarization model that incorporates both components of thermal emission and reflection, and a novel model of an LWIR polarimetric imaging system that accounts for the effects of stray light, camera characteristics, and thermal noise.
(2)~We present a physics-based learning method to estimate surface normals with a neural network trained on a synthetic dataset generated using the LWIR polarization model.
(3)~We implement a prototype system for demonstrations and construct \mbox{\thermalpol}, the first real-world benchmark dataset for LWIR SfP.
(4)~Experimental results demonstrate the effectiveness of our approach, achieving superior normal estimation accuracy compared to prior methods, particularly for transparent objects.


%% file: 02_related_work.tex
\section{Related Work}
\subsection{Shape from Polarization}
Shape from Polarization (SfP) is a classical yet actively evolving technique for estimating surface normals based on polarimetric measurements.  The fundamental principle of SfP lies in the relationship between the polarization state of light and the geometry of the surface it interacts with.  This relationship is exploited through either model-based approaches that leverage physical principles of polarization, or learning-based methods that extract geometric information from polarimetric data using neural networks. 

Model-based methods rely on the Fresnel equations to relate the degree of linear polarization (DoLP), denoted as $\dolp$ in this paper, to the zenith angle $\theta$ of the surface normal.  Similarly, the angle of linear polarization (AoLP), $\aolp$, is used to determine the azimuth angle $\phi$.  However, these mappings are fundamentally ambiguous. Specifically, there is a $\pi$-ambiguity due to the limited AoLP range of $[0,\pi]$, and a $\frac{\pi}{2}$-ambiguity arising from whether the reflection is diffuse or specular.
Regarding the zenith angle $\theta$, specular reflection poses a challenge due to ambiguity: the relationship with the DoLP $\dolp$ is non-monotonic, meaning the same DoLP value can map to two different angles around the Brewster angle. 

To resolve these ambiguities, various cues have been incorporated, including the direction of normals at object boundaries~\cite{miyazaki2003polarization}, shading information~\cite{mahmoud2012direct}, multi-spectral analysis~\cite{stolz2012shape,zhao2016multi,jinglei20183d,miyazaki2002determining}, multi-view imaging~\cite{miyazaki2003polarization2, miyazaki2017surface, cui2017polarimetric, tian2022high}, and fusion with depth maps~\cite{kadambi2017depth}.  Despite these efforts, such methods often struggle with scenarios involving a combination of diffuse and specular reflection, or non-uniform albedo.

In contrast, learning-based approaches~\cite{ba2020deep,lei2022shape} train the complex relationship between shape and polarization directly using neural networks.  This data-driven approach has demonstrated robustness to diverse materials and complex lighting conditions, overcoming the limitations of traditional model-based methods in non-ideal situations. Consequently, the applicability of SfP to a broader range of real-world scenarios has significantly improved.

However, both model-based and learning-based approaches face significant challenges when dealing with transparent objects, primarily due to the difficulty in observing surface reflections.  Existing techniques for transparent objects~\cite{miyazaki2002determining, miyazaki2003polarization2, shao2023transparent} employ a dome-shaped diffuse light source to create uniform reflections on the transparent surface. Nevertheless, these methods often neglect the complex internal reflections and transmission within the object, leading to inaccuracies.

\subsection{Long-Wave Infrared Imaging}

Long-wave infrared (LWIR) imaging, operating in the $8-$\SI{14}{\micro\meter} spectrum, has found widespread application in diverse fields, including thermography, night vision, and non-destructive testing~\cite{usamentiaga2014infrared}.  Recently, there is growing interest in expanding its use, offering potential in areas where visible-light techniques are insufficient.

Several studies have explored the use of LWIR for shape reconstruction applications. Tanaka~\etal~\cite{tanaka2019time} proposed a photometric stereo approach based on time-resolved light transport decomposition, but this method suffers from long acquisition times. Nagase~\etal~\cite{nagase2022shape} introduced a depth estimation technique exploiting wavelength-dependent LWIR attenuation, but the resulting depth maps often exhibit large errors. Narayanan~\etal~\cite{narayanan2024shape} presented an approach based on heat conduction, but this technique is restricted to thin, sheet-like objects.

Polarimetry in LWIR has been explored to various tasks, including reflection removal~\cite{li2018removal}, road surface recognition~\cite{li2021illumination}, and face recognition~\cite{hu2016polarimetric}. These studies focus on leveraging polarization cues for improved image understanding, not 3D shape reconstruction.

\subsection{Normal Estimation using LWIR Polarimetry}
Attempts to use LWIR for SfP date back to the 1990s, but they encountered significant hurdles.
Partridge~\etal~\cite{partridge1995three} demonstrated that the DoLP of emitted rays uniquely corresponds to the zenith angle of the surface normal and applied LWIR to SfP, although they reported large errors due to the camera noise. Miyazaki~\etal~\cite{miyazaki2002determining} argued that the capability of LWIR cameras at the time was insufficient for SfP and instead used LWIR polarimetry as an auxiliary cue to disambiguate zenith angles in visible light SfP. Kechiche~\etal~\cite{kechiche2017use} conducted experiments on materials such as glass and molten metal, but reported that the measured Stokes vectors did not appear to be physically valid, preventing accurate shape reconstruction.

While the limited performance of early sensors played a role, two fundamental problems, in particular, hindered accurate estimation in these prior works. (1) Neglect of reflection: In the LWIR spectrum, surrounding objects also act as light sources, and their reflections must be considered. (2) Neglect of systematic errors: Systematic errors in LWIR imaging, stemming from sensor's thermal noise and the stray light such as reflection/emission from the polarizer, were not adequately compensated.

In contrast, our work addresses these fundamental limitations by introducing a polarimetric model for the combined ray of emission and reflection, mitigating systematic errors by using a carefully calibrated imaging system, and further introducing a data-driven method to achieve high-precision normal estimation.

%% file: 03_principle.tex
\section{Principles}

\subsection{Polarization of Emission and Reflection}
\label{sec: model}
The polarization state of combined ray, whether reflected or emitted, is determined by the Fresnel equations, which govern the interaction of light at a surface. We treat emission as a process where unpolarized ray generated inside the material is transmitted through the surface and refracted as it emerges~\cite{wolff1998image}.
This emitted ray's polarization state is similar to that of diffuse reflection, where external incident ray first enters the material, scatters within it and then exits.
\begin{figure}[b]
    \centering
    \includegraphics[width=\columnwidth]{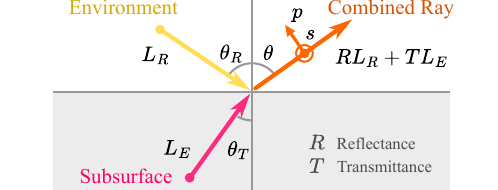}
    \caption{
        Radiance of a combined ray of the emitted component~$T L_E$ and the reflected component~$R L_R$.
    }
    \label{fig: fresnel}
\end{figure}

\begin{figure}
    \centering
    \includegraphics[width=0.8\columnwidth]{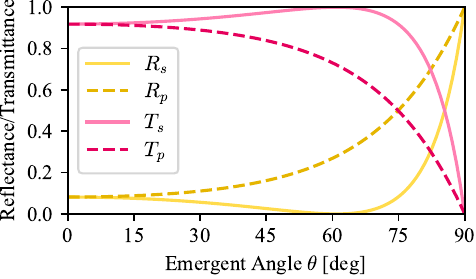}
    \caption{
        Reflectance and transmittance versus emergence angle $\theta$. Refractive index $\eta = 1.8$.
        For each of the $\mathit{p}$- and $\mathit{s}$-polarization components, the sum of reflectance and transmittance is always equal to 1.
    }
    \label{fig: ref_and_trans}
\end{figure}

To analyze the polarization state of the emergent ray, we consider the $\mathit{p}$- and $\mathit{s}$-polarization components for the reflectance and transmittance independently. As shown in~\cref{fig: fresnel}, for a given emergent angle $\zenith$, Snell's law relates the incident angles $\thetair$ and $\thetait$ for the reflected and transmitted rays, respectively:
\begin{align}
\thetair &= \zenith, \\
\eta \sin{\thetait} &= \sin{\zenith},
\end{align}
where $\eta$ represents the refractive index.
The reflectances for the $\mathit{p}$- and $\mathit{s}$-polarization components are given by the Fresnel equations:
\begin{align}
\refp &= \left( \frac{\tan(\thetair - \thetait)}{\tan(\thetair + \thetait)} \right)^2, \\
\refs &= \left( \frac{\sin(\thetair - \thetait)}{\sin(\thetair + \thetait)} \right)^2.
\label{eq: reftectance}
\end{align}
Similarly, the transmittances for the $\mathit{p}$- and $\mathit{s}$-polarization components are:
\begin{align}
\transp = 1 - \refp, \\
\transs = 1 - \refs,
\label{eq: transmittance}
\end{align}
which are derived from the principle of conservation of energy, as shown in \cref{fig: ref_and_trans}.

The $\mathit{p}$- and $\mathit{s}$-polarization components of the combined ray, $\out$, are expressed as a linear combination of the incident radiances:
\begin{align}
{\out}_p = \refp {\iref}_p + \transp {\itrans}_p, \\
{\out}_s = \refs {\iref}_s + \transs {\itrans}_s,
\end{align}
where $\iref$ and $\itrans$ denote the radiances of the incident rays from the external and internal sides of the surface, respectively.
They are expressed using the temperatures of the object and the surrounding $\temp_\textrm{obj}, \temp_\textrm{env}$ as
\begin{gather}
    \itrans = \stefan \temp_\textrm{obj}^4,
    \iref = \stefan \temp_\textrm{env}^4,
\end{gather}
where $\stefan$ is the Stefan-Boltzmann constant.
Assuming both $\iref$ and $\itrans$ are unpolarized, we have:
\begin{align}
{\out}_p = \frac{1}{2} (\refp \iref + \transp \itrans), \\
{\out}_s = \frac{1}{2} (\refs \iref + \transs \itrans).
\label{eq: radiance}
\end{align}
Finally, the DoLP $\dolp$ and the AoLP $\aolp$ are defined as:
\begin{gather}
\dolp = \left| \frac{{\out}_p - {\out}_s}{{\out}_p + {\out}_s} \right|, \label{eq: dolp}\\
\aolp =
    \begin{cases}
      \azimuth\pmod{\pi} & \text{if } {\out}_p > {\out}_s, \\
      \azimuth + \frac{\pi}{2}\pmod{\pi} & \text{if } {\out}_p < {\out}_s. 
    \end{cases} \label{eq: aolp}
\end{gather}
Since the AoLP is limited in the range of $[0,\pi]$, an inherent $\pi$ ambiguity exists.
Note that when ${\out}_p = {\out}_s$, the DoLP~$\dolp$, is zero, and the AoLP~$\aolp$, is undefined. 

\subsection{LWIR Polarimetric Imaging}
\begin{figure}
    \centering
    \includegraphics[width=0.95\columnwidth]{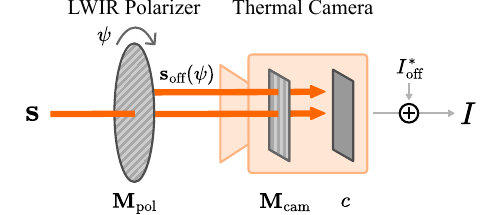}
    \caption{
        Our model of LWIR polarimetric imaging system.
    }
    \label{fig: imager}
\end{figure}

A LWIR polarimetric imaging system, which consist of a thermal camera and an LWIR polarizer, capture the polarization state of the ray emerged from the scene by acquiring images at multiple polarizer orientations (\cref{fig: imager}). Although this approach is analogous to polarimetric imaging in the visible spectrum, LWIR systems suffer from two major systematic errors that require careful correction, as noted by Eriksson~\textit{et al.}~\cite{eriksson2018polarimetric}. The first error is an offset component intrinsic to the imaging system, which is composed of thermal noise of the sensor and the stray light such as emission and reflection from the polarizer, both of which affect the recorded pixel values. The second error stems from sensor's polarization-dependency. Specifically, the surface structure of microbolometer causes the attenuation of certain polarization components~\cite{dem2011application}.

Taking these effects into account, we model the raw pixel value, $\iraw(\polangle, \ext)$, for the scene Stokes vector $\ext$ with a linear polarizer angle of $\polangle$ in front of the camera, as:
\begin{align}
    \iraw(\polangle, \ext) &= \gainvec^\top \mcam \mpol(\polangle) \ext + \offset,
    \label{eq: imaging_model}
\end{align}
where
\setlength{\jot}{7pt}
\begin{gather} 
    \gainvec = [\gain,0,0]^\top,\\ 
    \mcam =
    \frac{1}{2} \scalebox{0.7}{
        $\begin{bmatrix}
            1 + k & 1 - k & 0 \\
            1 - k & 1 + k & 0 \\
            0 & 0 & 2\sqrt{k}
        \end{bmatrix}$
    },\\
    \mpol(\polangle) =
    \frac{1}{2} \! \scalebox{0.7}{
        $\begin{bmatrix}
            1 &\!\! \cos(2\polangle) &\!\! \sin(2\polangle) \\
            \cos(2\polangle) &\!\!\!\! \cos^2(2\polangle) &\!\!\!\! \sin(2\polangle)\cos(2\polangle) \\
            \sin(2\polangle) &\!\!\!\! \sin(2\polangle)\cos(2\polangle) &\!\!\!\! \sin^2(2\polangle)
        \end{bmatrix}$
    }, \\
    \offset = \gainvec^\top \mcam \ext_\textrm{off}(\polangle) + I^{*}_\mathrm{off}.
\end{gather}
Here, we use a 3-dimensional Stokes vector and Mueller matrices to represent the polarization state, neglecting circular polarization. 
The scalar $\gain$ represents the camera gain and other scalar factors.
$\mcam$ represents the polarimetric response of the microbolometer, modeled as a partial linear polarizer. $k$ represents its gain factor at $\polangle = \SI{90}{\degree}$, while the gain factor is 1 at $\polangle = \SI{0}{\degree}$~\cite{collett2005polarizers}.
$\mpol(\polangle)$ is an ideal linear polarizer with an angle $\polangle$.
$\offset$ is the offset component. $\ext_\textrm{off}(\polangle)$ is the stray light such as emission/reflection by the polarizer, and $I^{*}_\mathrm{off}$ is the camera's thermal noise.


The scene's Stokes vector, $\ext$, can be calculated from measurements of $\iraw(\polangle, \ext)$ taken at multiple polarizer angles $\polangle$. This process requires $\gainvec, \mcam$ and $\offset$ to be known.

%% file: 04_method.tex
\section{Proposed Method}

\subsection{Reconstruct Stokes Vector from Images}
To reconstruct the scene Stokes vector $\ext$ from the captured pixel value $\iraw$, we need to determine three parameters in \cref{eq: imaging_model}: the gain $\gainvec$, the polarization response $\mcam$, and the offset component $\offset$.
The former two, $\gainvec$ and $\mcam$, are constant parameters and can be determined through prior calibration, though the offset $\offset$ varies over time due to the sensor temperature.
Eliminating this offset $\offset$ is crucial for both the calibration of $\gainvec$ and $\mcam$, and the measurement of the scene $\ext$.
We achieve this by taking the difference between two images. Assuming $\offset$ remains constant between the two captures, the difference image $I_{\Delta}$ eliminates the offset:
\begin{align}
    I_{\Delta}(\polangle, \ext_\beta, \ext_\alpha) = \gainvec^\top \mcam \mpol(\polangle) (\ext_\beta - \ext_\alpha).
    \label{eq: diff}
\end{align}

\begin{figure}[t]
    \includegraphics[width=1.0\linewidth]{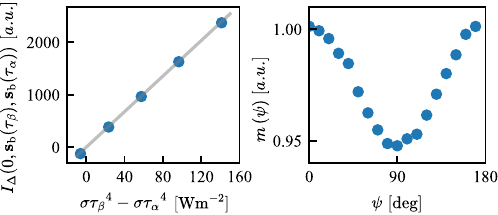}
    \caption{
    Characteristics of our imaging system.
    Left: Linear relationship between blackbody radiation differences and pixel intensity values across various temperature pairs ($\temp_\alpha$ = \SI{23}{\degreeCelsius}, $\temp_\beta$ = 20, 35, 50, 65, \SI{80}{\degreeCelsius}). This linearity validates our calibration approach based on differential measurements.
    Right: Polarization-dependent attenuation of the camera's microbolometer sensor, represented by gain factor $\mcampsi(\polangle)$ for rays with AoLP $\polangle$. The observed variation of approximately \SI{5}{\%} confirms the need for polarimetric correction in our imaging pipeline.
    }
    \label{fig: calib}
\end{figure}

For the calibration of $\gainvec$ and $\mcam$, we capture images of two blackbodies at different known temperatures and compute their difference.
According to the Stefan-Boltzmann law, an ideal blackbody at temperature $\temp$ emits thermal radiation, and its Stokes vector $\ext_\mathrm{b}(\temp)$ is given by:
\begin{equation}
    \ext_\mathrm{b}(\temp) = [\stefan \temp^4, 0, 0]^\top.
\end{equation}
Measuring difference images $I_{\Delta}(\polangle, \ext_\mathrm{b}(\temp_\beta), \ext_\mathrm{b}(\temp_\alpha))$ for several pairs of blackbody temperatures $(\temp_\alpha, \temp_\beta)$ enables the gain $\gainvec$ to be determined.
Similarly, by performing measurements at multiple polarizer angles $\polangle$, the polarization response matrix $\mcam$ can be determined~(\cref{fig: calib}).

\begin{figure*}[t]
    \centering
    \setlength{\unitlength}{1mm}
    \begin{picture}(160,38)
        \put(0,0){\includegraphics[width=1.0\linewidth]{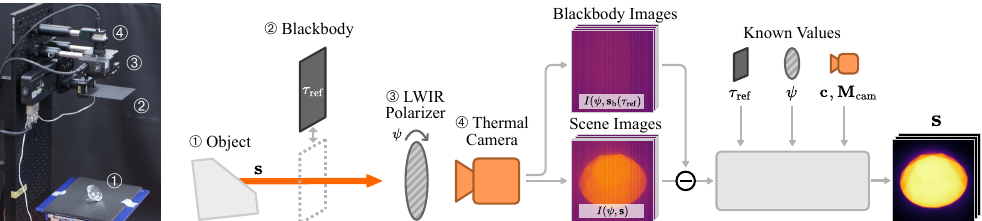}} 
        \put(123.7,5.6){\footnotesize \cref{eq: lsq}}
    \end{picture}
    \caption{
        LWIR polarimetric imaging system. Components: reference blackbody, rotating LWIR polarizer, and thermal camera.
        Scene Stokes vector $\ext$ is reconstructed from the difference between scene and blackbody measurements, eliminating system offsets while preserving polarization information.
    }
    \label{fig: system}
\end{figure*}

For measuring the Stokes vector $\ext$ from the scene, we compute the difference between images $I_{\Delta} \left(\polangle, \ext, \ext_\mathrm{b}(\temp_\mathrm{ref}) \right)$ of the scene and a reference blackbody at a known temperature $\temp_\mathrm{ref}$ (\cref{fig: system}).
Then we compute $\ext$ from multiple $I_{\Delta}$ measurements using the least-squares method:
\begin{align}
    \ext = (\mathbf{K}^\top \mathbf{K})^{-1} \mathbf{K}^\top \mathbf{I} + \ext_\mathrm{b}(\temp_\mathrm{ref})
    \label{eq: lsq}
\end{align}
where $\mathbf{I} \in \mathbb{R}^N$ and $\mathbf{K} \in \mathbb{R}^{N \times 3}$ are stacked $I_{\Delta}$ and $\gainvec^\top \mcam\, \mpol$, respectively, for each $\{\polangle_j\}_{j=1}^N$.


\subsection{Polarization-Normal Relationship in LWIR}
\label{sec: norm_and_pol}
In LWIR SfP, an important constraint exists: the temperatures of the object and its surroundings must be different. 
Since the AoLP of the emitted and reflected rays are orthogonal, their polarization components cancel each other out. 
When the intensities of these components ($\itrans$ and $\iref$) are equal, the DoLP becomes zero according to \cref{eq: dolp}, which disables obtaining surface normal information via polarimetry. 
To deal with this limitation, we opted for a simple solution: heating the object before capture so that the emitted component is dominant.

The relationship between the zenith angle $\zenith$ (equal to the emergent angle) and the DoLP $\dolp$, is illustrated in \cref{fig: zenith_and_dolp}~(right). 
Its profile differs significantly from that of the pure emission model, which was used in previous studies~\cite{partridge1995three, miyazaki2002determining, kechiche2017use}.
The combined ray exhibits a lower DoLP compared to the pure reflection or emission case.
Furthermore, similar to the pure reflection case, an ambiguity arises where a single $\dolp$ value can correspond to two distinct $\zenith$ values. However, this ambiguity is often not a significant issue in practice, as the DoLP peaks at a larger $\zenith$ values. Under the configuration of \cref{fig: zenith_and_dolp}, the DoLP reaches its peak at a zenith angle of approximately $\zenith \approx \SI{79}{\degree}$. Importantly, over a large portion of the object's surface (corresponding to \SI{98.2}{\%} of the projected radius for a sphere), the DoLP increases monotonically with $\zenith$, as depicted in \cref{fig: zenith_and_dolp}~(bottom). 

The azimuth angle is equal to the AoLP, $\aolp = \azimuth$, because the emission component is dominant.

\subsection{Model-based Estimation}
\begin{figure}[t]
    \centering
    \includegraphics[width=1.0\columnwidth]{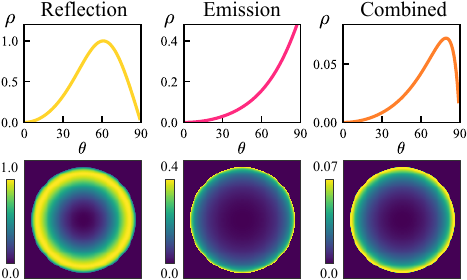}
    \caption{
        Relationships of the emergent angle and the DoLP.
        Top: Degree of polarization $\dolp$ corresponding to the normal zenith angle $\zenith$. Refractive index $\eta = 1.8$. In combined ray scenario, $\iref : \itrans = 1:0.7$, which is obtained from Stefan-Boltzmann's law assuming that the object temperature and environmental temperature are \SI{50}{\degreeCelsius} and \SI{23}{\degreeCelsius}, respectively.
        Bottom: DoLP map when imaging a sphere.
    }
    \label{fig: zenith_and_dolp}
    \vspace{-2ex}
\end{figure}

Based on the observation in~\cref{sec: norm_and_pol}, we propose a model-based normal estimation method. To determine the zenith angle, we utilize the relationship shown in \cref{fig: zenith_and_dolp}, obtaining the solution by assuming that the zenith angle is less than the peak.
We assume the refractive index $\eta$ and the emission-reflection radiance ratio $L_R:L_T$ are known, then resolve the $\pi$ ambiguity in azimuth by iteratively determine the azimuth angle, propagating inward from the object boundary while enforcing smoothness in the resulting normal field, under the assumption that the normal at the object's boundary points outward, following Miyazaki~\textit{et~al.}~\cite{miyazaki2003polarization}.

\subsection{Learning-based Estimation}
We propose a learning-based method to complement model-based approaches, which are susceptible to intricate geometries and noise. In this approach, we generate a synthetic dataset through simulating our polarization physics models, and train a neural network.
We employs a UNet-based architecture~\cite{ronneberger2015u}, augmented with Transformer blocks to capture the global image context crucial for resolving the azimuth ambiguity in SfP. This global context is particularly crucial in LWIR imaging, where shading cues are absent.

Following Lei~\etal~\cite{lei2022shape}, we insert Transformer blocks between the encoder and decoder~\cite{chen2021transunet, yang2021transformer}, applying sinusoidal positional encoding to effectively utilize global context. This network has fewer parameters compared to the original UNet~\cite{ronneberger2015u} and its variants designed for SfP~\cite{kondo2020accurate, ba2020deep, lei2022shape}. Even a smaller network can achieve high accuracy because of the simple optical characteristics in LWIR.

The input tensor is a concatenation of four polarization images ($L_{0}, L_{\frac{1}{4} \pi}, L_{\frac{1}{2} \pi}, L_{\frac{3}{4} \pi}$) and polarization information ($L, \dolp, \cos{2 \aolp}, \sin{2 \aolp}$). 
The polarization image for $\polangle$ is defined as:
\begin{equation}
    L_\polangle = \frac{1}{2} (s_0 + s_1 \cos{2 \polangle} + s_2 \sin{2 \polangle})
\end{equation}
where $[s_0, s_1, s_2]^\top = \ext$. The AoLP is encoded using its trigonometric components to handle angular continuity, following Lei~\textit{et~al.}~\cite{lei2022shape}. The object mask is then applied to this tensor.

The output is the surface normals represented in view-vector space. In this coordinate system, the per-pixel z-axis is aligned with the view-vector. Unlike camera space, view-vector space changes from pixel to pixel in a perspective camera. This view-vector coordinate system is well-suited for representing the relationship between surface normals and the view-vector, as the DoLP $\dolp$ directly corresponds to the angle between them. The output normals are then transformed into camera space based on the camera's focal length.

Our network is trained on a synthetic dataset generated by simulating our model, using cosine similarity loss. Further implementation details are presented in \cref{sec: training_process}.

\begin{figure}[t]
    \centering
    \includegraphics[width=\columnwidth]{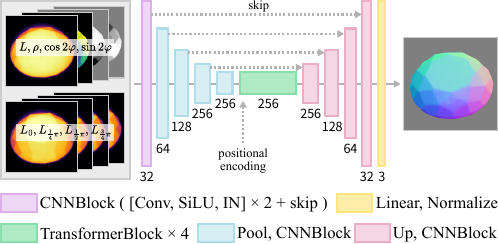}
    \caption{
        Our neural network, based on UNet~\cite{ronneberger2015u}, features CNN encoder and decoder, employing SiLU activation functions~\cite{elfwing2018sigmoid} and instance normalization~\cite{ulyanov2016instance}, with Transformer modules~\cite{vaswani2017attention, dosovitskiy2020image} as a bottleneck.
    }
    \label{fig: nn}
\end{figure}

%% file: 05_implementation.tex
\section{Implementation}
\subsection{Imaging System and Real-world Dataset}
\subsubsection{Imaging System Setup}
\label{sec: implementation_setup}
Our LWIR imaging system, shown in \cref{fig: system}~(left), employs an uncooled microbolometer thermal camera (FLIR Boson: \SI{12}{bit}, 640$\times$512 pixels, \SI{60}{fps}, \SI{18}{\degree} HFoV) and a holographic wire-grid polarizer (Thorlabs WP50H-B). The polarizer is mounted on a rotation stage (OptoSigma: OSMS-60YAW), and rotates during scene acquisition. After capturing the scene, a blackbody board is placed in front of the camera, and a second set of images is acquired. This board is an aluminum plate coated with blackbody spray and equipped with a thermal sensor. Our prototype system requires one second for scene capture and approximately seven seconds total, including the blackbody board's movement and subsequent image acquisition. 

\subsubsection{Real-world Dataset}
For quantitative evaluation, we introduce \thermalpolsixteen, a real-world dataset comprising LWIR polarimetric images with corresponding ground-truth shapes. This dataset encompasses 16 objects of diverse materials and shapes, including glass, ceramics and plastics. In the data acquisition process, the objects were heated to approximately \SI{50}{\degreeCelsius} within a heating cabinet, while the ambient temperature was maintained at roughly $21-$\SI{23}{\degreeCelsius}. To mitigate reflections from high-temperature sources, such as humans and ceiling lights, the imaging system was shielded with a cardboard enclosure. The object shapes were obtained using a scanner (Revopoint MINI 2) after coating the objects with white spray paint, and ground truth normal maps were generated by aligning the 3D models with the LWIR images using MeshLab~\cite{meshlab}.

\subsection{Synthetic Dataset and Training Process}

\subsubsection{Synthetic Dataset}
Our network was trained on a synthetic dataset generated using polarized rendering.
We expect this to allow the neural network to handle phenomena such as inter-reflection, except for extreme cases.
We utilized the Mitsuba3~\cite{mitsuba3} for this process. Since Mitsuba3 lacks support for polarized emission from surfaces, we developed a custom material that simulates it. This material is implemented as a surface with injection of uniform, unpolarized rays from behind.

While this simulation faithfully represents the physical model, it may have biases compared to real-world conditions, particularly because it assumes a fixed refractive index and a smooth surface.

The training dataset comprises 86 images: 43 unique 3D models captured from two distinct viewpoints. During training, data augmentation is employed: each image is rotated by 36 distinct angles, randomly cropped, and random Gaussian noise added. Throughout the dataset generation, the refractive index was held constant at 1.8, while the ratio of ambient light intensity to the simulated emitted light intensity was randomly varied between 0.6 and 0.7 for each image.

\subsubsection{Training Process}
\label{sec: training_process}
The network is optimized using Adam~\cite{kingma2014adam} optimizer with parameters $\beta_1 = 0.9$, $\beta_2 = 0.999$, and $\epsilon = 10^{-8}$. We train the network for 100 epochs with a batch size of 8. The learning rate is initialized to $10^{-4}$ and subsequently decayed by a factor of 0.5 every 10 epochs.

\begin{figure}[t]
    \centering
    \includegraphics[width=1.0\linewidth]{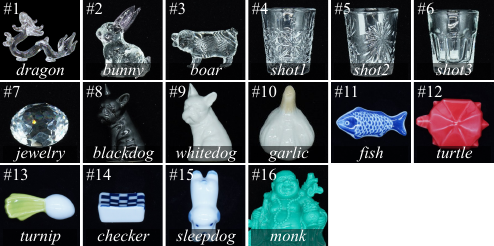}
    \caption{
        Objects in \thermalpolsixteen dataset.
    }
    \label{fig: real_dataset}
\end{figure}

\begin{figure}[t]
    \centering
    \includegraphics[width=1.0\linewidth]{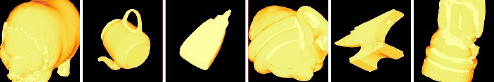}
    \caption{
        Examples of synthetic training dataset.
    }
    \label{fig: synth_dataset}
\end{figure}

%% file: 06_experiment.tex
\section{Experiments}
\subsection{Evaluation on \thermalpol Dataset}
First, we evaluate the accuracy and robustness of the proposed methods on the \thermalpolsixteen dataset, which includes the ground truth data. 
\cref{table: thermalpol} shows the mean angular errors (MAE) of normals estimated by our model-based (MB) and learning-based (LB) methods on the \thermalpolsixteen dataset. The results demonstrate high accuracy, with total average MAEs of approximately \SI{10}{\degree} for most objects, including transparent ones, particularly with the learning-based method.

\begin{table}[t]
\centering
\caption{Mean angular errors [\si{\degree}] on \thermalpol.}
\begin{tabular*}{0.98\linewidth}{@{\extracolsep{\fill}}lcccccc}
\toprule
\tablerow{{},\#1,\#2,\#3,\#4,\#5,\#6}
\cmidrule(){2-7}
\tablerow{MB,14.0,18.2,12.1,14.8,14.7,14.5}
\tablerow{LB,13.9,13.2,11.9,9.7,9.4,8.3}
\midrule
\tablerow{{},\#7,\#8,\#9,\#10,\#11,\#12}
\cmidrule(){2-7}
\tablerow{MB,12.6,10.5,11.5,7.3,8.8,9.5}
\tablerow{LB,9.9,7.6,8.3,7.1,7.7,9.9}
\midrule
\tablerow{{},\#13,\#14,\#15,\#16,\multicolumn{2}{c}{mean}}
\cmidrule(){2-7}
\tablerow{MB,11.6,10.3,8.7,36.4,\multicolumn{2}{c}{13.5}}
\tablerow{LB,7.4,9.2,7.8,23.4,\multicolumn{2}{c}{10.3}}
\bottomrule
\end{tabular*}
\label{table: thermalpol}
\end{table}

\subsubsection{Comparisons between Model-based and Learning-based Methods}
\begin{figure*}[t]
    \centering
    \includegraphics[width=1.0\linewidth]{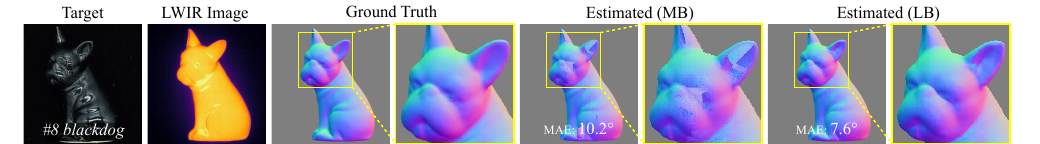}
    \caption{Comparisons between model-based and learning-based methods.}
    \label{fig: pb_vs_lb}
    \vspace{3ex}
    \centering
    \includegraphics[width=1.0\linewidth]{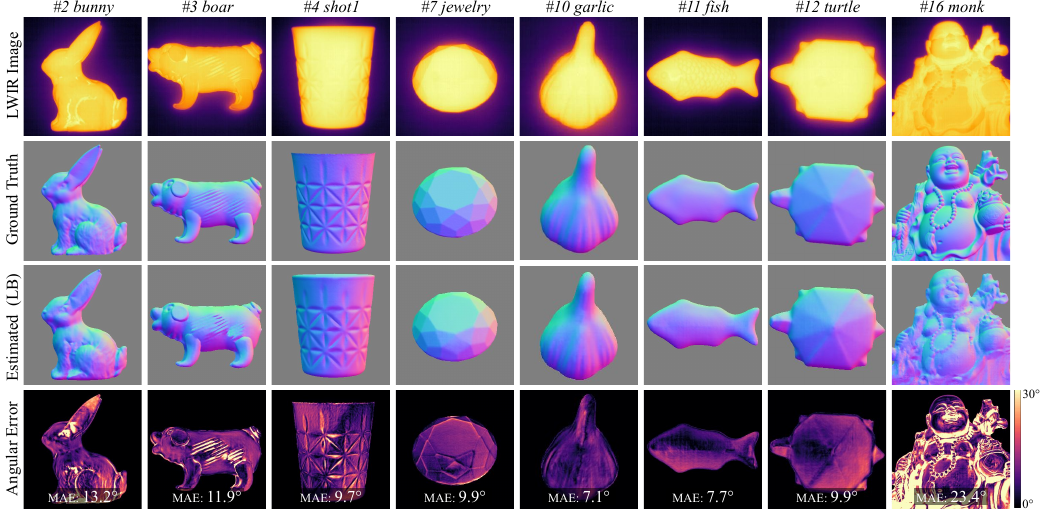}
    \caption{Quantitative evaluation of the learning-based method on the \thermalpol dataset.}
    \label{fig: results}
\end{figure*}

\Cref{fig: pb_vs_lb} presents a comparison between normal maps estimated using the model-based and the learning-based methods for the \blackdog object. Both approaches effectively approximate ground truth normals across substantial portions of the object's surface, though the model-based method exhibits a higher MAE. This performance gap stems primarily from two factors: inaccurate azimuth angle estimation resulting from limitations in the propagation algorithm, and noise in regions with small zenith angles where low DoLP values make measurements particularly vulnerable to camera noise. It is worth emphasizing that the learning-based method, despite being trained exclusively on synthetic data, demonstrates superior generalization to the real-world dataset compared to the model-based approach.

\subsubsection{Qualitative Evaluation}
\cref{fig: results} presents the normals estimated by the learning-based method alongside their corresponding angular error maps.
Normals across substantial portions of the surfaces are estimated with low error. In particular, objects with simpler geometries, such as \garlic and \fish, show excellent reconstruction quality. However, significant errors are evident in the convex crease regions of more complex objects like \boar and \monk. These errors primarily stem from inter-reflections, which manifest in exceptionally bright regions of the LWIR images where smooth surfaces specularly reflect LWIR radiation from other parts of the objects. In these regions, the combination of emission and reflection causes a substantial reduction in DoLP, significantly compromising our ability to extract reliable shape information.

\subsection{Additional Experiments}
\begin{figure}[t]
    \centering
    \includegraphics[width=1.0\linewidth]{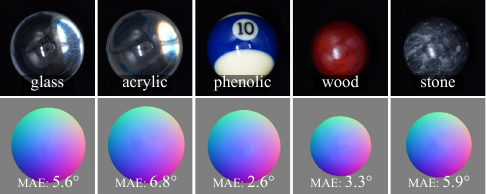}
    \caption{Evaluation of the impact of material properties on the proposed method.}
    \label{fig: spheres}
\end{figure}

\subsubsection{Impact of Materials}
We comprehensively evaluate the impact of material properties on our normal estimation approach.
\cref{fig: spheres} shows the estimated normals for a diverse set of spheres fabricated from glass, acrylic, phenolic, wood, and stone.
The LWIR polarization technique demonstrates a significant advantage by successfully reconstructing object shapes regardless of their optical properties in the visible spectrum, including transparency, translucency, and surface albedo. While all materials yield reasonable shape estimates, we observe that estimation accuracy varies across different materials, likely due to material-specific differences in refractive index and emissivity that influence the polarization characteristics.

\begin{figure}[t]
    \centering
    \includegraphics[width=1.0\linewidth]{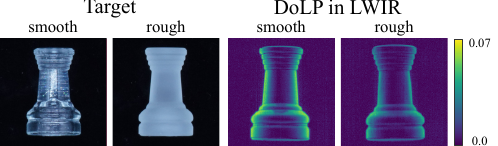}
    \caption{DoLP of glass objects with both the smooth and rough surfaces.}
    \label{fig: roughness}
    \vspace{3ex}
\end{figure}
\subsubsection{Effect of Surface Roughness}
We also evaluate the effect of surface roughness on the polarization state. \cref{fig: roughness} illustrates the DoLP for both smooth and rough glass objects. The measurements reveal that rough surfaces consistently exhibit lower DoLP values compared to smooth surfaces, despite the generally diminished influence of micro-facets at longer wavelengths. This finding demonstrates that surface roughness remains a significant factor affecting polarization state even in the LWIR spectrum, potentially leading to inaccurate zenith angle estimation when using LWIR SfP techniques.

\subsubsection{Cooled Object}
To evaluate the versatility of our approach across varying thermal conditions, we conduct an experiment where an object is cooled prior to measurement, in contrast to our standard heated-object protocol. \cref{fig: hotcool} demonstrates the significant differences in polarization states between heated and cooled objects. Most notably, the relationship between surface normal and polarization parameters changes substantially, with the AoLP and azimuth angle exhibiting a $\pi/2$ \si{rad} difference in the cooled condition. This angular shift occurs because cooling the object increases the dominance of the reflected component in the polarization signal. Impotantly, our network--trained on a synthetic dataset that included simulations of objects with little emission--successfully estimated surface normals from LWIR polarimetric images of the cooled object, demonstrating the robustness of our approach across different thermal scenarios.

\subsubsection{Performance in Uncontrolled Environments}
\begin{figure}[t]
    \centering
    \includegraphics[width=1.0\linewidth]{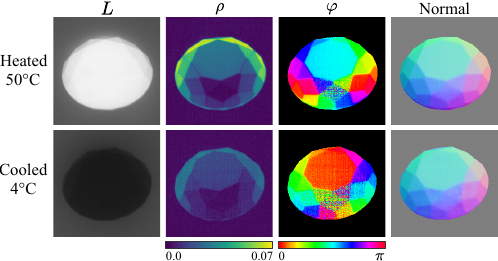}
    \caption{
    Polarization states and estimated normal maps for a heated/cooled object, with the surroundings at \SI{21}{\degreeCelsius} in both cases.
    }
    \label{fig: hotcool}
    \vspace{3ex}
    \centering
    \includegraphics[width=\columnwidth]{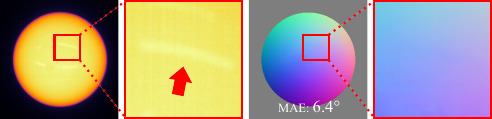}
    \caption{
        Example of a glass sphere captured without an enclosure (left: $\ext_0$, right: normal map).
        Despite the visible reflections from the ceiling lights, the neural network handled them correctly.
    }
\label{fig: uncontrolled}
\end{figure}

We evaluate our method's performance under non-uniform environmental a condition.
\cref{fig: uncontrolled} illustrates a result from an indoor setting without the enclosure.
While the assumption of a uniform environment is generally valid indoors due to the minimal temperature variations between objects, it can be compromised by reflections from localized hot spots.
Our learning-based approach is designed to mitigate such artifacts, as they are similar to the inter-reflections it already accounts for. 
\cref{fig: uncontrolled} demonstrates effective handling of these challenging reflections with our method. 
Nevertheless, further analysis across more diverse uncontrolled environments is required.

\subsubsection{Comparisons across Different Network Architectures}
\begin{table*}[t!]
\centering
\caption{Comparisons between network architectures trained on \thermalpolsixteen.}
\begin{tabular}{l *{9}{c}}
    \toprule
     & \multicolumn{3}{c}{Angular Error [°] ↓} & \multicolumn{3}{c}{Accuracy [\%] ↑} & \multicolumn{2}{c}{Training costs ↓} & \multicolumn{1}{c}{Inference cost ↓} \\
    \addlinespace[0.7ex]
     & {Mean} & {Median} & {RMSE} & {11.25°} & {22.5°} & {30°} & {Params [M]} & {VRAM [MB]} & {Time [ms]} \\
    \midrule
    \tablerow{DeepSfP\cite{ba2020deep},11.34,9.14,14.92,68.24,92.91,95.33,13.3,206,14.9}
    \tablerow{SfPWild\cite{lei2022shape},11.52,9.54,14.64,64.11,92.67,95.80,42.5,701,35.7}    \tablerow{Kondo~\textit{et~al.}\cite{kondo2020accurate},11.04,9.11,13.72,66.27,92.97,95.83,19.5,334,14.1}
    \tablerow{UNet\cite{ronneberger2015u},10.93,9.39,13.43,64.83,93.63,96.42,31.1,507,21.4}
    \tablerow{Ours (MB),13.46,10.60,17.94,61.28,87.78,91.43,\text{--},\text{--},636.8}
    \tablerow{Ours (LB),\tbf{10.28},\tbf{8.96},\tbf{12.57},\tbf{69.16},\tbf{94.78},\tbf{96.71},\tbf{6.6},\tbf{148},\tbf{11.6}}
    \bottomrule
\end{tabular}
\label{table: networks}
\end{table*}
We conducted a comprehensive evaluation of various network architectures for our learning-based method using the \thermalpol dataset. We trained established SfP networks; DeepSfP~\cite{ba2020deep}, SfPWild~\cite{lei2022shape}, and Kondo~\etal~\cite{kondo2020accurate}, along with the original UNet~\cite{ronneberger2015u} on our synthetic dataset. Our evaluation metrics included mean, median, and RMSE of angular error, as well as accuracy (percentage of pixels with error below threshold). Additionally, we compared practical implementation factors such as number of learnable parameters, training VRAM requirements, and per-image inference time. For reference, we included results from our model-based method.

Our proposed network achieves superior performance with the lowest angular errors and highest accuracy among all compared methods, while requiring less than half the parameters of competing architectures (\cref{table: networks}). Notably, other architectures, including the original UNet, also demonstrate robust performance. This finding presents an interesting contrast to previous studies in visible light SfP~\cite{lei2022shape, shao2023transparent}, which reported limited effectiveness from networks not specifically designed for SfP applications. Our results therefore suggest that LWIR polarimetry provides inherently rich information for normal estimation and exhibits greater robustness to architectural choices compared to visible light SfP.

\subsubsection{Effect of the Number of Images on Stokes Vector Reconstruction}
\begin{figure}[t]
    \centering
    \includegraphics[width=1.0\linewidth]{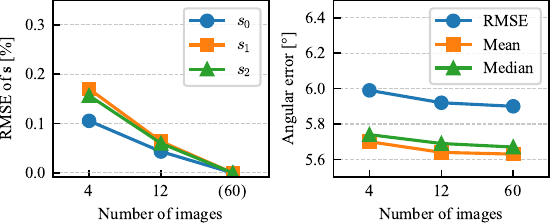}
    \caption{
    Evaluation of Stokes Vector Reconstruction with fewer images. The Stokes parameters $(s_0,s_1,s_2)$ were reconstructed using varying number of images $(4, 12, 30, 60)$ and compared against the $60$-image reconstruction as reference. Target object is \textit{glass sphere}.
    Left: Errors of Stokes Vector. Normalized relative to $s_0$ pixel-wisely and then root-mean-squared across the entire object region.
    Right: Angular Error of estimated normals. 
    }
    \label{fig: numimg}
\end{figure}

We analyzed how the number of acquired images affects the quality of reconstructed $\ext$ values, leveraging our prototype's capability to capture $60$ images in a single acquisition (\cref{fig: numimg}). As expected, Stokes vectors reconstructed from fewer images exhibit increased noise. However, the quantitative difference between $\ext$ values reconstructed using just $4$ images versus the full $60$ images is remarkably small, merely $0.1-$\SI{0.15}{\%}. Normal estimation accuracy remained robust, with the 4-image reconstruction achieving a MAE of only \SI{5.7}{\degree}, comparable to using all $60$ images. 
This finding suggests that our normal estimation method can perform effectively even with minimal image acquisition, offering potential benefits for computational efficiency and reducing capture time.

%% file: 07_conclusion.tex
\section{Conclusion}
We presented a novel LWIR SfP approach that effectively handles transparent objects by explicitly modeling both thermal emission and reflection. Our method addresses limitations of previous techniques through: (1) a comprehensive polarization model accounting for emission and reflection, (2) an accurate LWIR polarimetric imaging model compensating for systematic errors, and (3) a physics-based learning approach trained on synthetic data.

We implemented a prototype system and created ThermoPol, the first real-world benchmark dataset for LWIR SfP. Experiments validated our method's high accuracy across various materials, including those transparent to visible light. Our learning-based approach outperforms traditional model-based methods with mean angular errors of approximately \SI{10}{\degree} while requiring fewer parameters than competing architectures.

Limitations include: (1) a several-second acquisition time, potentially addressable through pixelated LWIR polarizers~\cite{chenault2016pyxis}; (2) requirement for temperature difference between object and surroundings; and (3) challenges with materials having substantial variations in emissivity/refractive index or highly rough surfaces.

Despite these limitations, our work demonstrates the significant potential of LWIR polarimetry for shape reconstruction of transparent objects that challenge traditional computer vision techniques, opening new possibilities for passive shape acquisition across a broader range of materials and applications.


